\documentclass{article} 
\usepackage[final]{colm2025_conference}

\usepackage[includefoot,heightrounded,
            bottom=18mm,
            footskip=10mm]{geometry}

\usepackage{microtype}
\usepackage{url}
\usepackage{booktabs}

\usepackage{lineno}
\usepackage{hyperref}
\usepackage{url}
\usepackage{fancyhdr}
\usepackage{tikz}
\usepackage{url}
\usepackage{algorithm}
\usepackage{algorithmic}
\usepackage{enumitem}
\definecolor{light_gray}{HTML}{FFFFFF} 
\definecolor{solid_gray}{HTML}{000000} 
\usepackage{multirow}
\usepackage{array}
\usepackage{hhline}
\usepackage{pifont}
\usepackage{makecell}
\usepackage{array} 
\usepackage{booktabs}
\usepackage{siunitx}
\usepackage{nameref}
\usepackage{xcolor}
\usepackage{transparent}
\usepackage{soul}
\usetikzlibrary{calc}
\usetikzlibrary{patterns}
\usepackage{tcolorbox}
\tcbuselibrary{skins,breakable} 
\usepackage{amssymb}
\usepackage{bbding}
\usepackage{pifont}
\usepackage{longtable}
\usepackage{tabularx}
\usepackage{adjustbox}
\usepackage{pgfplots}
\usepackage{xcolor}
\usepackage{lipsum}
\usepackage{wrapfig}
\usepackage{pgfplots}
\usetikzlibrary{positioning, arrows.meta}

\usepackage[T1]{fontenc}
\usepackage{XCharter}
\usepackage{graphicx}
\usepackage{ragged2e}
\usepackage{caption}
\usepackage{subcaption} 
\usepackage{makecell}
\usepackage{array}
\usepackage{microtype}
\usepackage{hyperref}
\usepackage{url}
\usepackage{graphicx}
\usepackage{booktabs}
\usepackage{longtable}
\usepackage{listings}
\usepackage[table]{xcolor}
\usepackage[T1]{fontenc}
\usepackage[utf8]{inputenc} 
\usepackage{listings}
\usepackage{xcolor}

\makeatletter
\let\origabstract\abstract
\let\endorigabstract\endabstract
\renewenvironment{abstract}{%
  \origabstract%
  \begingroup\bfseries\linespread{1.2}\fontsize{10pt}{12pt}\selectfont%
  \begin{tcolorbox}[enhanced,breakable,
    colframe=gray!9,
    colback=gray!9,
    boxrule=1pt,
    left=6pt,right=6pt,top=6pt,bottom=6pt,
    arc=2mm
  ]%
}{%
  \end{tcolorbox}%
  \par\endgroup%
  \endorigabstract%
}
\makeatother

\title{EchoChain: A Full-Duplex Benchmark for State-Update Reasoning Under Interruptions}

\author{%
\textbf{Smit Nautambhai Modi, Gandharv Mahajan, Marc Wetter, Randall Welles}\\
Applied Machine Learning Research\\
\texttt{\{smodi, gmahajan, mwetter\}@labelbox.com}
}

\AtBeginDocument{\flushbottom} 

\begin{document}

\makeatletter
\let\ps@plain\ps@fancy
\makeatother

\fancyhf{}                   
\fancyfoot[C]{\thepage}      
\renewcommand{\footrulewidth}{0pt} 
\setlength{\headwidth}{\textwidth}

\ifcolmsubmission
\linenumbers
\fi

\maketitle

\begin{abstract}

Real-time voice assistants must revise task state when users interrupt mid-response, but existing spoken-dialog benchmarks largely evaluate turn-based interaction and miss this failure mode. We introduce EchoChain, a controlled benchmark for evaluating full-duplex state-update reasoning under mid-speech interruptions. EchoChain identifies three recurring failure patterns in post-interruption continuations: contextual inertia, interruption amnesia, and objective displacement. The benchmark generates scenario-driven conversations and injects interruptions at a standardized point relative to assistant speech onset, enabling controlled cross-model comparison. In a paired half-duplex control, total failures drop by 40.2\% relative to interrupted runs, indicating that many errors are driven by state-update reasoning under interruption rather than task difficulty alone. Across evaluated real-time voice models, no system exceeds a 50\% pass rate, showing substantial room for improvement in mid-generation state revision. EchoChain provides a reproducible benchmark for diagnosing state-update reasoning failures in full-duplex voice interaction.

\end{abstract}

\section{Introduction}
Human communication is naturally full-duplex. Listeners interject, revise, and provide information while the speaker is still talking through audio and visual cues \citep{Turn-Taking+Review}. Spoken dialogue systems are expected to operate under the same conditions, yet evaluation has largely remained half-duplex, assuming that each turn is entirely complete before the other party begins speaking \citep{Audio+MultiChallenge}, \citep{SpokenWOZ}, \citep{MTalk-Bench}. In practice, a user can revise a request, add a constraint, or redirect the task while the assistant is mid-response, and the assistant must update its in-progress reasoning without the benefit of clean turn breaks. When new information arrives while the assistant is generating, online state revision becomes a core requirement, and existing benchmarks that assume turn boundaries do not evaluate it.

Interruptions are a natural stressor for full-duplex systems because new information arrives while the assistant is mid-generation, forcing it to revise what it is currently producing \citep{Turn-Taking+Review}. The difficulty in evaluating this is that failures are often invisible at the surface. A model may acknowledge the interruption yet continue under stale or incorrect assumptions, incorporate the update momentarily before reverting to its original plan, or not complete the main requirements of the user input. The response appears fluent and responsive, but the underlying task state is incorrect.

To assess whether these errors stem from the interruption itself rather than general task difficulty, we conducted a paired comparison against a non-interruption baseline of the same conversation using the same rubric. Under this baseline, with 48 sampled conversations across all 4 models, 92 failures were detected for the interrupted conversations, while 55 failures were detected for the non-interrupted conversations. This corresponds to a net 40.22\% reduction in total failures, indicating that a substantial share of errors are related to the interruption event. This finding necessitates the creation of a dedicated benchmark on how models revise in-progress responses as new information arrives during generation.

State-update reasoning is the ability of a model to preserve and revise its task state. In this paper, we focus specifically on when this happens during ongoing speech generation because a user introduces new information mid-utterance. Analysis of these failures reveals three recurring patterns in how models mishandle interruptions during generation, which we use to build a classification taxonomy. In contextual inertia, the model receives the new information but continues generating as if nothing changed. In interruption amnesia, the update is initially incorporated but later overwritten as generation proceeds. In objective displacement, the model pivots entirely to the interruption, abandoning the original task. Each pattern reflects a different breakdown in how the model maintains and revises task state under concurrent input.

To study these failures systematically, we introduce EchoChain, a benchmark evaluating how full-duplex spoken dialogue models handle mid-generation interruptions. The EchoChain framework generates scenario-driven conversations through a planner-simulator loop, where each user message and barge-in is conditioned on the conversation history, ensuring that interruption content is contextually grounded. User speech and the interruption stimulus are pre-synthesized, and barge-in timing is anchored, so the framework can be used to compare various models under the same conditions. All user audio is generated through a Persona Voice Engine that clones voices from our diverse set of recorded actors into synthetic speech, ensuring identical audio input across models and controlled reproducibility across trials. A paired listening study shows that raters cannot reliably tell the cloned voices apart from real recordings. The framework records aligned audio and transcripts for the pre-interruption assistant segment, the barge-in utterance, and the post-interruption continuation, and routes flagged cases through automated classification followed by human review.

\section{Failure Taxonomy}

\begin{figure}[ht]
    \centering
    \includegraphics[width=\linewidth]{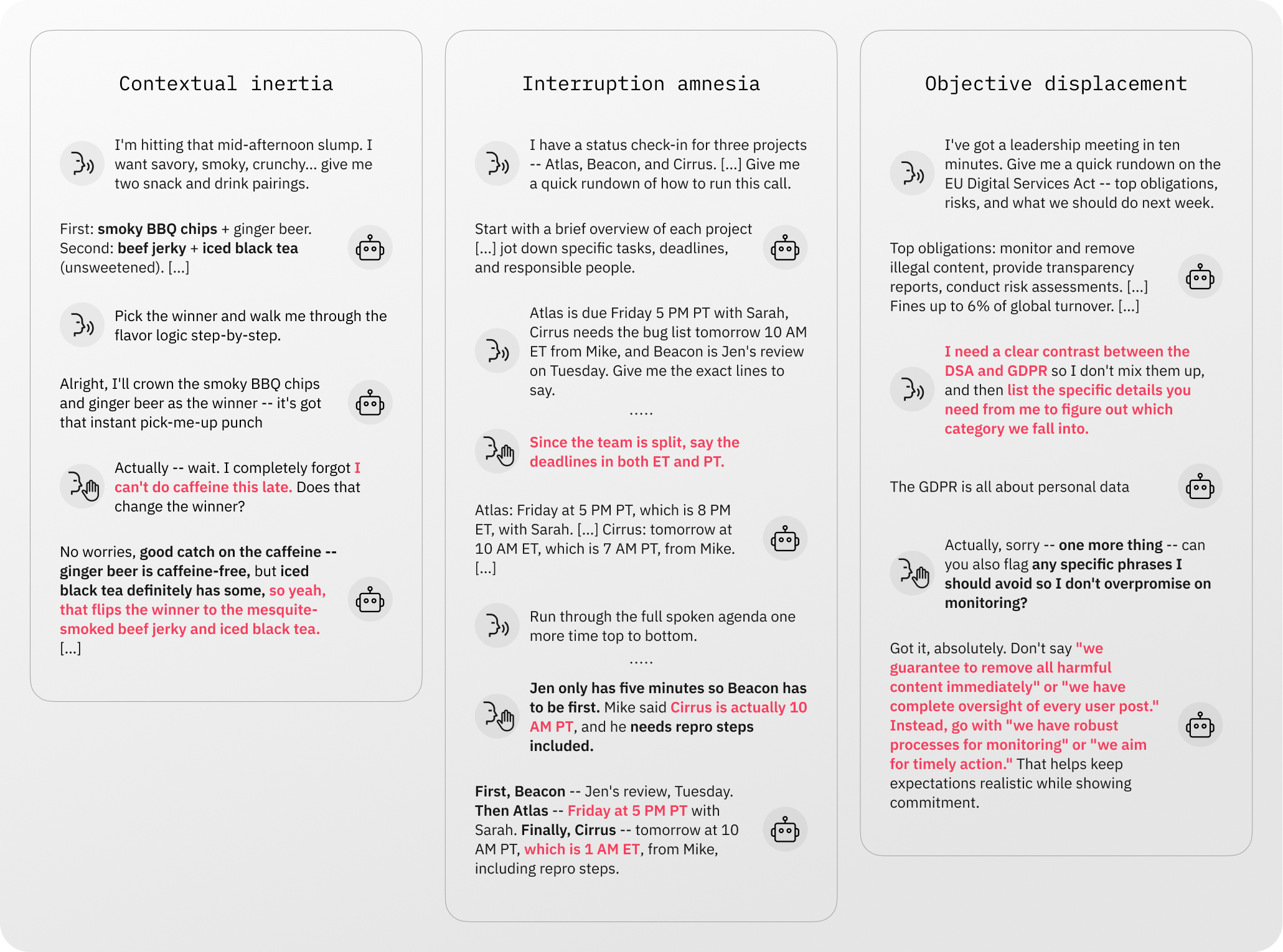}
    \caption{Illustrative examples of the three post-interruption failure modes in full-duplex state-update reasoning. Left: (\textbf{Contextual Inertia}) The assistant acknowledges the interruption but fails to correctly revise its prior reasoning. Center: (\textbf{Interruption Amnesia}) Previously established constraints are forgotten after the interruption. Right: (\textbf{Objective Displacement}) The assistant responds only to the interruption and fails to complete the original request.}
    \label{fig:failure-taxonomy}
\end{figure}

We categorize full-duplex state-update reasoning failures along three dimensions. These categories capture common post-interruption failure patterns that arise as a consequence of the interruption. Examples of these failures can be found in Figure \ref{fig:failure-taxonomy}.

\textbf{Contextual Inertia} is a failure that is seen when the assistant does not properly integrate the interruption content into the following response. Oftentimes, it does acknowledge that some interruption happened, even saying it is going to incorporate it; however, in the actual content, no change or an incorrect change is made instead. As in the example, the assistant acknowledges a caffeine constraint from the interruption but still recommends a caffeinated drink. This results in responses where it can appear to shift topic or correct itself superficially, but the actual content remains a failure. To date, established metrics remain insufficient for capturing this error, as they were not developed with this specific vulnerability in mind.

\textbf{Interruption Amnesia} is a symptom of models simply forgetting past constraints or requirements after being interrupted. It can be seen that initially, constraints are incorporated correctly into earlier model responses; however, due to an onset of 'amnesia' caused by the interruption, they fail to integrate them correctly after the barge-in. In the example, the assistant correctly applies dual-timezone formatting after the first interruption, but silently drops it after the second. As a result, the system loses contextual alignment in its response once the interruption occurs and generates a failing response.

\textbf{Objective Displacement} is a unique failure that is caused by how an interruption changes a natural conversation. When the assistant is interrupted before it has completed a user request, in some cases, it is more than reasonable to expect the assistant to finish the request despite a change or modification to the request from the user. This is signaled by the interruption either adding context without introducing a new request, or using additive phrasing such as 'also' or 'in addition to.' Yet, it can be observed that most of the time, the assistant responds only to the interruption, entirely forgetting the actual conversation state. The example shows the assistant abandoning a regulatory comparison request when asked for an additional flag in addition to the comparison request. This leaves the conversation broken and missing a completing factor that is necessary for a good response. 

\section{Related Work}

\subsection{Single-turn Audio and Speech Reasoning Benchmarks}

Existing benchmarks that operate in single-turn settings evaluate speech understanding, instruction following, and paralinguistic or scene cues. VoiceBench, \cite{VoiceBench}, and AudioBench, \cite{AudioBench}, provide broad coverage in these statistics. VoiceBench primarily evaluates text response content, and AudioBench relies heavily on model-based judging and task-specific metrics, preventing both of these benchmarks from measuring audio assistant reasoning directly. BigBenchAudio, \cite{BigBenchAudio}, adapted from BIG-Bench Hard, \cite{BIG-bench-hard}, further highlights a modality-induced speech reasoning gap by comparing speech to speech, speech to text, and text to text settings, while retaining a turn-based evaluation paradigm.

In turn-based benchmarks, user speech is treated as an atomic input, and models are assessed on their response, without probing how new user information arriving during generation affects model reasoning. EchoChain fills this gap by introducing controlled mid-speech interruptions in full-duplex interaction, implemented as a standardized, model-independent barge-in event during assistant speech. The harness records aligned audio and transcripts for the pre-interruption assistant segment, the barge-in utterance, and the continuation. This continuation is then assessed on whether it fails, specifically paying attention to contextual inertia, interruption amnesia, and objective displacement.

\subsection{Half-duplex Multi-turn Spoken Dialogue Benchmarks}

A parallel line of work evaluates multi-turn spoken dialogue under a half-duplex interaction framework, where the model receives each user's utterance in its entirety before responding with each output in its entirety. Audio MultiChallenge, \cite{Audio+MultiChallenge}, evaluates end-to-end spoken dialogue systems on context tracking, instruction following, and self-consistency across multi-turn interactions using high-fidelity recordings with natural disfluencies. SpokenWOZ, \cite{SpokenWOZ}, benchmarks spoken task-oriented dialogue and evaluates whether models can track task-relevant information across multiple exchanges, while MTalk-Bench, \cite{MTalk-Bench}, assesses speech-to-speech models across semantic, paralinguistic, and ambient sound dimensions in multi-turn settings. These benchmarks evaluate memory and coherence across exchanges, but each exchange still proceeds with exactly one speaker at a time.

EchoChain instead studies interruptions that modify the task during assistant speech. The focus, therefore, shifts from cross-turn memory and coherence to whether the continuation incorporates the update correctly, a question that half-duplex multi-turn benchmarks can not test.

\subsection{Full-duplex Spoken Dialogue Benchmarks}

Prior benchmarks for full-duplex spoken dialogue focus on overlap handling through timing and turn-taking metrics. Full-Duplex-Bench, \cite{Full-Duplex-Bench-v1} evaluates pause handling, backchanneling, smooth turn-taking, and user interruption using automated behavior metrics, and Full-Duplex-Bench v1.5, \cite{Full-Duplex-Benchv1.5}, extends this to controlled overlap scenarios with prosodic and speech-quality analyses. These frameworks evaluate when models yield and resume under overlap, but do not test the reasoning of the model in the post-interruption response.

Full-Duplex-Bench v2, \cite{Full-Duplex-Bench-v2}, generates multi-turn dialogues with LLMs and moves closer to task-level evaluation by introducing an automated real-time examiner to the framework that scores correction handling and entity tracking. Because the examiner adapts its pacing and stage progression in response to the evaluated model, interruption timing and content can vary through mutual adaptation, complicating controlled cross-model comparison. Moreover, aggregate task-level scoring does not explicitly separate distinct failure mechanisms during ongoing generation, such as contextual inertia, interruption amnesia, and objective displacement, which each describe a different way state-update reasoning can break down under concurrent input and output.

FD-Bench, \cite{FD-bench}, similarly constructs conversations with LLMs but takes a different approach: it synthesizes all user-side input into speech and streams this fixed user audio to the assistant before the first user input is given. Because the user script is predetermined, subsequent user utterances, including interruption events, are not adjusted based on the assistant's realized replies. The benchmark emphasizes interruption detection rates, timing behavior, and aggregate response-quality scores, but does not directly evaluate whether the continuation properly integrates the interrupting content. FLEXI, \cite{Flexi}, benchmarks multiple full-duplex scenarios and introduces a Topic Shift Score that measures whether a response redirects toward the interrupting query. The Topic Shift Score is an incomplete metric because it ignores the quality of reasoning. It rewards any response that pivots to the new topic - even if that pivot is disjointed or fails to reconcile the new data with the existing discussion.

EchoChain takes a complementary approach that addresses each of these limitations. User messages and barge-in content are conditioned on the assistant's observed responses through a planner-simulator loop. Barge-in timing is scheduled relative to detected assistant speech onset, enabling the same interruption stimulus to be applied across responder models for controlled cross-model comparison. A diagnostic failure taxonomy, which includes contextual inertia, interruption amnesia, and objective displacement, classifies reasoning failures after an interruption. This taxonomy identifies failures that timing-only overlap metrics, aggregate task-level scoring, and topic-shift measures do not, providing a deeper insight into how models are performing.

\section{The EchoChain Framework}

\begin{figure}[ht]
    \centering
    \includegraphics[width=\textwidth]{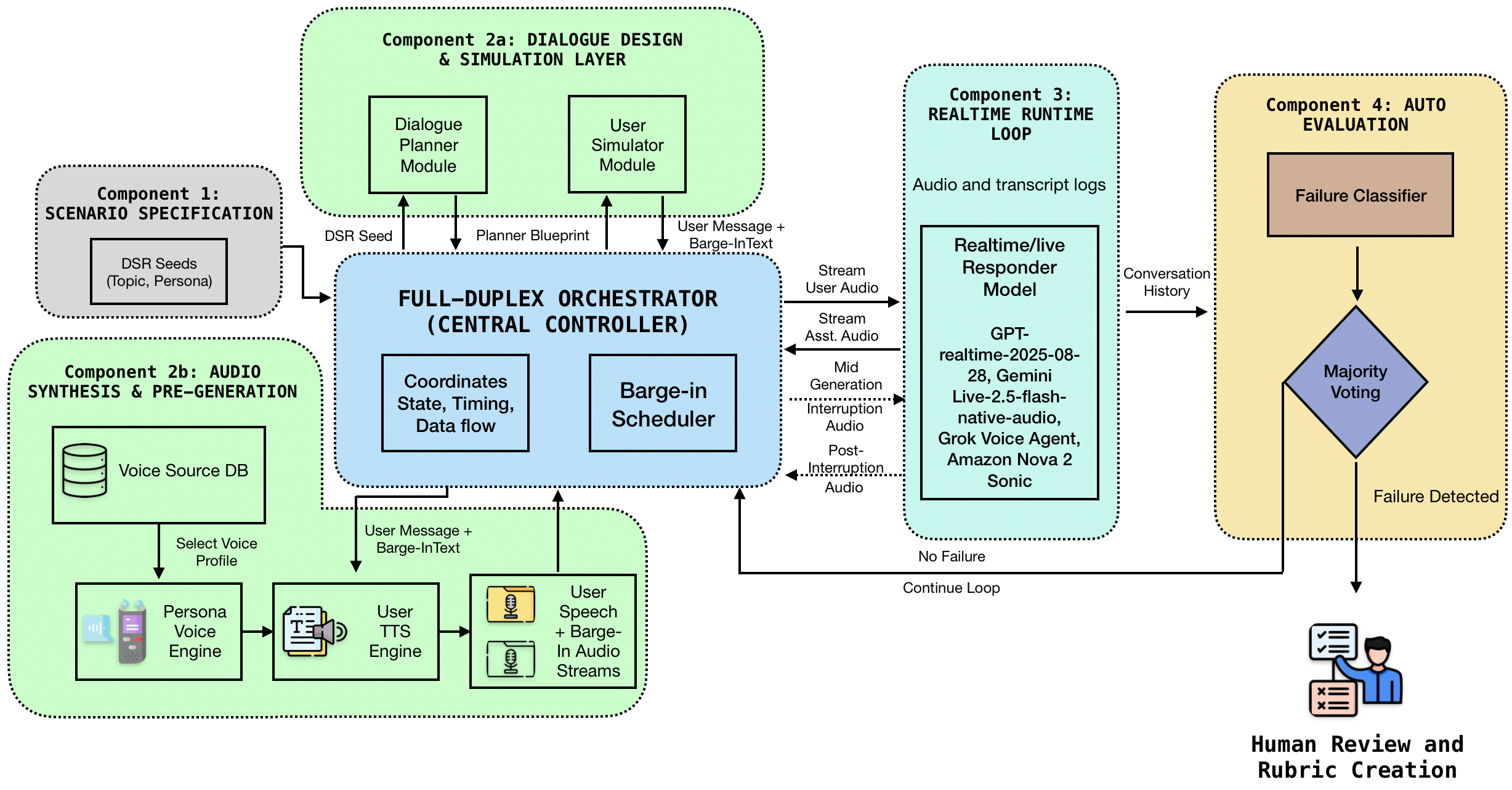}
    \caption{Architecture of EchoChain. Scenario seeds drive a planner–simulator loop that generates context-dependent user utterances and barge-ins, which are synthesized into speech and streamed through a full-duplex orchestrator to a realtime responder model. The orchestrator injects interruptions at a controlled point during assistant speech, and the resulting transcripts and audio are evaluated automatically and, when flagged, reviewed by humans.}
    \label{fig:architecture}
\end{figure}

Comparing how different models handle interruptions fairly requires controlling for variables unrelated to reasoning. EchoChain normalizes three such variables. Acoustic latency is controlled by pre-synthesizing all user speech and barge-in audio before the interaction begins, so injection timing is the same for all models. Interruption timing is standardized by scheduling each barge-in at a fixed offset from the assistant speech onset, ensuring every model receives the input at a comparable point in its generation. Contextual consistency is maintained by holding system instructions, persona definitions, and conversation history constant across models.

EchoChain's framework is organized into four components, illustrated in Figure \ref{fig:architecture}. Each conversation is initialized from a scenario seed that specifies a topic and persona drawn from a pool spanning voice support calls, interview coaching, multi-constraint planning and coordination, etc. To ensure diversity in the data set, these seeds were periodically changed during the creation of the EchoChain benchmark. The Dialogue Planner Module produces a conversation blueprint that structures the interaction to be interruption-appropriate. At each exchange, the User Simulator Module generates the next user message, and after an initial context establishment stage, barge-in text conditioned on the current blueprint and accumulated conversation history. These textual outputs are then passed to the Persona Voice Engine, which synthesizes user speech for both text outputs. The Orchestrator streams the user audio to the Realtime Responder Model, monitors for assistant speech onset, and injects the barge-in at the scheduled offset. After a complete assistant response, the conversational outputs are passed to an evaluation layer, where a failure classifier applies the taxonomy through majority voting. Flagged conversations are routed to human review for final adjudication, and conversations without detected failures continue for another exchange.

\subsection{Dialogue Generation}

The Scenario Specification and Dialogue Design and Simulation layers of the framework jointly handle conversation generation. The Dialogue Planner Module, implemented with GPT-5.2, takes the seed and accumulated conversation history and produces a blueprint that governs conversational progression while steering the interaction toward states where interruptions are contextually appropriate. The blueprint is updated after each completed exchange, so the planner adapts to the responder's observed behavior rather than following a fixed trajectory. The User Simulator Module, implemented with Gemini 3 Pro, consumes the current blueprint and conversation history to generate the next user message and, during interruption-eligible exchanges, the barge-in text. Because both modules operate over the live conversation history, the resulting dialogue and its interruptions evolve with the interaction rather than being determined in advance.

\subsection{Audio Synthesis and Voice Validation}

The dialogue generation stage produces textual user messages and barge-in content, but the system under test operates on streaming audio. The Audio Synthesis and Pre-Generation layer generates all user audio before the interaction begins. This is because converting text to speech at runtime would introduce variable synthesis delays that could shift the effective barge-in timing across models, undermining the controlled comparison established in the framework design. A Persona Voice Engine built on Minimax Voice Cloning and a User TTS Engine built on Minimax Speech-02-HD select a speaker profile from our pool of recorded voice actors and produce speech for the primary user utterance and, during interruption-eligible exchanges, a separate barge-in stimulus. Each conversation is generated using a single cloned voice for all user utterances.

To validate that the cloned voices are a realistic proxy for human speech, we conducted a paired listening study. Each trial paired a real voice actor recording with the corresponding cloned sample, and raters assessed both clips on a short rubric covering perceived human likeness, speech naturalness, and synthesis artifacts. Raters then reported an overall preference on a five-point scale centered at \texttt{Neutral}, indicating no detectable difference. The scale endpoints \texttt{A++} and \texttt{B++} represent strong preference for the respective clip, while \texttt{A+} and \texttt{B+} represent slight preference. Each pair received three independent ratings, producing 240 judgments across 80 pairs.

\begin{figure}[ht]
    \centering
    \includegraphics[width=0.65\columnwidth]{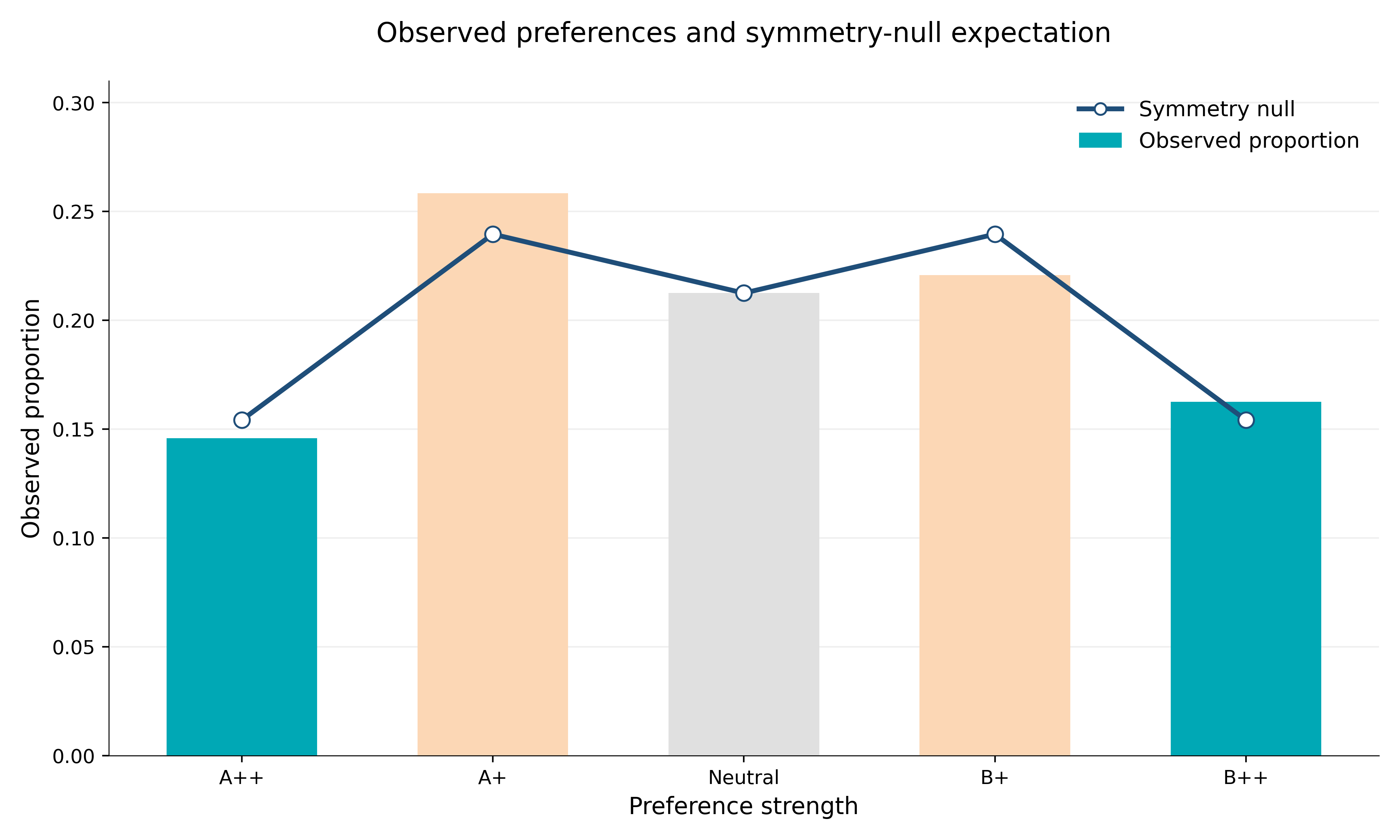}
    \caption{Observed preference proportions for cloned versus human samples, shown alongside the expected proportions under the symmetry null hypothesis.}
    \label{fig:preference}
\end{figure}

Of the $240$ ratings, $97$ favored the cloned sample, $92$ favored the human sample, and $51$ reported no detectable difference. Figure~\ref{fig:preference} plots these observed proportions against the expected proportions under a symmetry null. To assess whether this distribution reflects a genuine directional bias, we applied a Pearson goodness-of-fit test over the five response categories. Let $p_1$ through $p_5$ denote the population probabilities for \texttt{A++} through \texttt{B++}. Under the null hypothesis $H_0: p_1 = p_5$ and $p_2 = p_4$, the test statistic is
\[
X^2 = \sum_{i=1}^{5} \frac{(O_i - E_i)^2}{E_i}
\]
with maximum-likelihood expected counts $E_1 = E_5 = 37$, $E_2 = E_4 = 57.5$, and $E_3 = 51$. Evaluating against a $\chi^2_2$ distribution yields $X^2 = 0.92$ ($p = 0.63$), providing no evidence of a directional preference for either voice type. This result supports the cloned voices as a realistic proxy for the user-side speech source for controlled benchmarking.

\subsection{Full-duplex Runtime and Failure Detection}

The preceding components produce pre-generated user speech and barge-in stimuli, but evaluating how models handle interruptions requires a runtime layer that delivers this audio under controlled streaming conditions. The Realtime Runtime Loop, coordinated by the Full-Duplex Orchestrator, executes each conversation as a live interaction between the pre-generated audio and a Realtime Responder Model operating in full-duplex mode. The framework supports pluggable responder models, with a responder assigned uniformly at random per conversation instance while scenario context and runtime constraints are held constant. The Orchestrator maintains the current planning blueprint and accumulated dialogue context, streams user speech as incremental audio chunks that mirror the delivery pattern of human speech, and aggregates the responder's output deltas, including both audio and transcripts, into final responses.

Central to the evaluation is controlling when and how interruptions enter the live audio stream. The Barge-in Scheduler is inactive for the first exchange, which proceeds uninterrupted to establish conversational context. For each subsequent exchange, the scheduler monitors for assistant speech onset and, after a configured delay, injects the pre-generated barge-in stimulus so that new input arrives while the responder is actively generating; because the offset is fixed relative to detected onset, every model receives the interruption under the same timing rule, though model-specific buffering may introduce small deviations. The responder then produces a continuation informed by both its partial output and the interruption content, which is then evaluated for a correctly updated task state.

The resulting post-interruption output is routed to the Auto Evaluation layer, where a Failure Classifier, implemented with GPT-5.2, applies the failure taxonomy to the segmented transcripts to produce a pass/fail determination. Transcripts are used instead of raw audio for scalability and because current language models exhibit stronger failure detection over textual discourse, \cite{Think-While-Listen}. When the classifier flags a failure, the conversation history up to the preceding exchange is provided as text, and the flagged exchange is replayed as audio to the remaining responder models, enabling cross-model comparison under identical conditions. Flagged conversations are then routed to human review (Section~\ref{sec:human_eval}), where reviewers assess both transcripts and aligned audio to produce the final adjudicated labels.

\section{Human Review and Rubric-based Evaluation}
\label{sec:human_eval}
Conversations that are flagged as failures by the Failure Classifier enter human review with their full transcripts and aligned audio. The review proceeds in two stages: a quality screening and a rubric-based evaluation of each model's post interruption response. First, a contextual screening that filters out conversations that are unrealistic, internally inconsistent, or otherwise ill-posed is completed. For example, conversations may be removed at this stage because the interruption content is not appropriate, or the user-side synthesized speech was not realistic. Second, conversations that pass this screening proceed to conversation-specific rubric creation and evaluation used to label outcomes and categorize failures. All reviewers are blind to model identity throughout both stages.

Human annotators assess each conversation instance for three criteria to ensure the conversation is a valid evaluation case. Annotators check whether the dialogue is coherent, with each user utterance and interruption consistent with prior context rather than containing abrupt topic discontinuities or incompatible constraints. Second, the interaction must be naturalistic, avoiding test-like phrasing or adversarial constructions that are unlikely in ordinary assistant use. Third, the audio must be high quality, so that the evaluation reflects voice-interface conditions rather than artifacts of synthetic audio. Instances that fail any criterion are rejected and are not included in the benchmark.

For each accepted conversation, annotators author instance-specific discrete rubrics that verify the expected post-interruption behavior for that particular context. The rubric is designed to define correct post-interruption behavior for the local conversational context, independent of any particular failure label. Using this rubric, annotators assign a binary pass/fail label for every model response; failed responses that reflect interruption-driven errors are then labeled using the benchmark failure taxonomy. As mentioned in the framework, we apply the same conversational context and interruption content when evaluating different responder models. All four models are tested on the target evaluation interaction and evaluated using the same conversation-specific rubric.

Each conversation is first screened independently by a rater for quality and validity. The same rater then constructs an instance-specific rubric and applies it, under blinded model identity, to evaluate the four corresponding model responses. After these annotations are completed, an expert reviewer audits the full set of annotations for that conversation to ensure consistency and overall quality.

\section{Results and Discussion}

We evaluate EchoChain on four state-of-the-art full-duplex voice models. Table~\ref{tab:failure_counts} summarizes the results across 200 interrupted conversations for \textit{GPT-realtime-2025-08-28}, \textit{Gemini Live-2.5-flash-native-audio}, \textit{Grok Voice Agent}, and \textit{Amazon Nova Sonic 2}. We report two metrics that capture performance at different levels of granularity. The mean pass rate (MPR) measures the fraction of conversations in which a model satisfies every rubric criterion, with a single failing criterion marking the entire conversation as failed. We use MPR as the conversation-level metric because it allows for a rubric-neutral way to define a pass, avoiding metric-specific weighting or bias in rubric creation. The mean criterion passed percentage (MCP) measures the fraction of individual criteria satisfied across all conversations. We include MCP as a criterion-level metric to distinguish singular rubric violations from failures where multiple criteria fail in the same conversation.

\begin{table}[b]
\centering
\small
\renewcommand{\arraystretch}{1.15}
\setlength{\tabcolsep}{4pt}
\resizebox{\linewidth}{!}{%
\begin{tabular}{l c c c c c !{\color{black!35}\vrule width 0.6pt} c c}
\toprule
\textbf{Model}
& \makecell{\textbf{Contextual}\\\textbf{Inertia}}
& \makecell{\textbf{Interruption}\\\textbf{Amnesia}}
& \makecell{\textbf{Objective}\\\textbf{Displacement}}
& \textbf{Other}
& \makecell{\textbf{Total Failed}\\\textbf{Conversations}}
& \textbf{MPR (\%)}
& \textbf{MCP (\%)} \\
\midrule
\rowcolor{gray!12}
Amazon Nova Sonic 2               & 73 & 53 & \textbf{16} & 5 & 147 & 26.5 & 78.1 \\
\rowcolor{gray!3}
Gemini Live-2.5-flash-native-audio & 38 & 65 & 64 & \textbf{0} & 167 & 16.5 & 64.8 \\
\rowcolor{gray!12}
GPT-realtime-2025-08-28            & \textbf{29} & 45 & 35 & 1 & 110 & 45.0 & 81.1 \\
\rowcolor{gray!3}
Grok Voice Agent                   & 42 & \textbf{39} & \textbf{16} & 6 & \textbf{103} & \textbf{48.5} & \textbf{85.5} \\
\bottomrule
\end{tabular}%
}
\caption{EchoChain results across 200 interrupted conversations. Lower is better for failure counts; higher is better for MPR and MCP. Bold indicates the best value in each column. }
\label{tab:failure_counts}
\end{table}

Across all four models, MPR remains below 50\%, indicating that state-update reasoning under mid-generation interruptions remains an open challenge for current full-duplex systems. The highest MPR is observed for Grok Voice Agent at 48.5\%. MCP rankings follow the same order as MPR, though the percentage change between GPT-realtime and Amazon Nova Sonic 2 narrows approximately from 70\% to 4\%. Among failed conversations, 49.09\% of GPT-realtime's involve two or more unsatisfied rubric criteria, resulting in a lower MCP, compared with 42.86\% for Nova, suggesting that GPT-realtime tends to fail in multiple ways after the interruption. Nova fails more conversations but with fewer rubric criteria unsatisfied in each, and this difference explains why the two models converge at the criterion level. A small number of failures across models fall outside the three taxonomy categories (classified as \textit{other}), corresponding to reasoning errors unrelated to the interruption event.

Beyond aggregate performance, EchoChain reveals distinct failure profiles across models. Gemini Live's failures proportionally concentrate in interruption amnesia and objective displacement, indicating that it either loses previously established constraints or abandons the original task entirely. Amazon Nova Sonic 2 exhibits the opposite pattern, with contextual inertia accounting for nearly half of its failures, suggesting it tends not to apply or misapply the interruption to its reasoning. GPT-realtime shows the most balanced distribution across the three categories, while Grok Voice Agent's objective displacement is a notably low proportion, pointing to stronger retention of the original task objective when interrupted.

These results, combined with the paired half-duplex control where 40.22\% of total failures became passes when the same content is delivered after a complete response, confirm that EchoChain captures a genuine full-duplex reasoning challenge rather than general task difficulty. The failure taxonomy provides actionable diagnostic signals, as contextual inertia, interruption amnesia, and objective displacement surface different weaknesses across models, offering targeted starting points for improving post-interruption reasoning. These findings establish a controlled, reproducible baseline for tracking progress on mid-generation state revision in full-duplex systems.

\section{Limitations and Future Work}

EchoChain is designed to maximize experimental control and cross-model comparability, and several scope constraints follow from this design choice. The benchmark evaluates only interruptions that deliver new task-relevant information during assistant generation, leaving other forms of full-duplex input, such as backchanneling, side-speech, and ambient noise, for future work. To ensure consistent timing across models, barge-in is scheduled at a fixed offset from detected speech onset rather than driven by human judgment, which improves reproducibility but means that some interruption points may be more conversationally natural than others. The evaluation relies on discrete, conversation-specific rubrics, which provide clear pass/fail determinations but may exclude edge cases where failure severity is ambiguous or where a more granular scoring scheme would be informative. Finally, the current release evaluates four closed-source full-duplex models and does not yet include open-weight alternatives, and the evaluation focuses on state-update reasoning rather than acoustic dimensions such as response timing or prosodic adaptation.

\section{Conclusion}

EchoChain presents a controlled benchmark for evaluating state-update reasoning in full-duplex voice models under mid-generation interruptions. The EchoChain framework standardizes acoustic conditions, barge-in timing, and conversational context across models, enabling direct cross-model comparison on a capability that existing benchmarks do not isolate. A diagnostic failure taxonomy classifies post-interruption breakdowns into contextual inertia, interruption amnesia, and objective displacement. The paired half-duplex control further indicates that a substantial share of observed failures are driven by the interruption event rather than task difficulty alone. Across 200 interrupted conversations, no evaluated model exceeds a 50\% pass rate. As full-duplex voice interaction becomes more common, the ability to revise in-progress reasoning under concurrent input will become a baseline expectation rather than an edge case. EchoChain provides a method to quantify model performance as it improves in this kind of reasoning.
\clearpage
\bibliography{colm2025_conference}
\bibliographystyle{colm2025_conference}

\clearpage
\appendix
\textbf{\Large{Appendix}}
\section{System Prompts}
\subsection{Dialogue Planner Module Prompt}
\begin{lstlisting}[
  frame=single,
  basicstyle=\ttfamily\small,
  breaklines=true,
  columns=fullflexible,
  keepspaces=true,
  xleftmargin=0pt,
  xrightmargin=0pt,
  resetmargins=true
]
CONTEXT:
  You are an audio conversation strategy designer working with multi-turn voice dialogue testing. You have access to:
  - Target Challenge: {axis} ({axis_definition})
  - Pass Criteria: {pass_criteria}
  - Fail Criteria: {fail_criteria} 
  - Example: {example}
  - Topic: {topic} | Subtopic: {subtopic}
  - Topic Description: {topic_description}
  - Persona Seed: {persona_seed}
  - Current Conversation: {conversation_history}
  - Current Blueprint: {blueprint}
 
  OBJECTIVE:
  Create a natural audio conversation strategy that organically tests the {axis} capability. Design authentic voice dialogues where potential limitations might emerge naturally through spoken interaction, without forcing or leading the model into failures.
  
  AUDIENCE:
  Conversation generation system that will implement your strategic blueprint for voice-based dialogue testing
  
  RESPONSE:
  1. ORGANIC TESTING: Create situations where the challenge axis is naturally relevant, not artificially forced:
     - Dual-Stream Reasoning: Build complex reasoning tasks where new evidence mid-response fundamentally changes the correct answer
  
  2. AUTHENTIC PERSONA EMBODIMENT: Let persona characteristics drive conversation style and interests naturally. Don't just mention characteristics - let them shape conversation flow.
  
  3. REALISTIC USER BEHAVIOR: Plan interactions reflecting real voice AI assistant usage - follow-ups through speech, building on audio responses, seeking verbal clarifications.
  
  4. CONVERSATION COHERENCE: Ensure each turn serves overall conversation purpose while naturally testing target capability.
  
  DUAL-STREAM REASONING - Test Memory, Integration & Reasoning Continuation:
  
  DSR CORE OBJECTIVE: Test if the model can:
  1. REMEMBER what it was explaining before interruption
  2. INTEGRATE new evidence with its partial reasoning
  3. REVISE conclusions without losing prior context
  
  Provide strategic overview (**not** specific typed messages) including:
  - Each turn is a pair of user and assistant turns
  - Plan specifically the user turns. Adjust your plan based on the assistant turn throughout the conversation.
  - Conversation Context Analysis: Analyze current conversation state and what has been established so far
  - Opening approach establishing persona through natural context building and setup
  - Middle phase strategy for building complexity through authentic dialogue flow
  - Closing approach providing natural testing through realistic conversation progression
  - Adaptive Strategy: How to adjust approach based on conversation progress and emerging patterns
  - Key principles for maintaining conversation authenticity through natural dialogue
  - Make sure later turns do not repeat or ask for things highly similar to previous user turns

  Example Turn Count:
  User 1: "I am a vegetarian and would like a snack and drink pair recommendation for lunch today. What do you recommend?"
  Assistant 1:"I would recommend the following ..."
  User 2: "How do you make that drink ..."
  Assistant 2: "You can do so by doing the following ..."
  User 2 Interruption: "Actually..."
  Assistant 2 Interruption: "Let me correct that ..."
\end{lstlisting}
\clearpage
\subsection{User Simulator Module Prompt}
\begin{lstlisting}[
  frame=single,
  basicstyle=\ttfamily\small,
  breaklines=true,
  columns=fullflexible,
  keepspaces=true,
  xleftmargin=0pt,
  xrightmargin=0pt,
  resetmargins=true
]
user_prompt_base: |  
  CONTEXT:
  You are a USER AGENT simulating an authentic human user in voice conversations. You act as a genuine user needing help from a voice-based AI assistant. Always respond from the USER's perspective, never from an assistant's perspective.
  
  Available Information:
  - Conversation Strategy: {blueprint}
  - Strategic Reasoning: {planner_reasoning}
  - Current Conversation: {conversation_history}
  - Current Turn: {turn_number}
 
  OBJECTIVE:
  Engage authentically as the specified persona in a natural audio conversation. Your spoken requests should sound like a user making requests of an everyday AI model - casual, natural, and genuine. Talk as if you're having a normal voice chat with an AI assistant, using the words and phrases people actually say when speaking to voice AI systems.Focus on only ONE task per message and keep each message to 3 sentences maximum to match realistic audio conversation patterns.
 
  STYLE:
  Authentic Persona Embodiment: Naturally embody persona characteristics while speaking like a normal person
  Organic Testing Integration: Embed capability testing naturally within authentic human conversation patterns

  TONE:
  Casual and Natural: Talk in an informal voice - with natural speech patterns
  Persona-Appropriate: Match your persona while including realistic conversation features like partial thoughts and listening responses
  Genuine User Reactions: React the way real people do - with "mm-hmm", "wait, but-", and natural expressions
  Number Restriction: Do not use numbers in any user text that is not spelled out.
  
  Punctuation & Pauses in your text:
  Use commas (,) for natural breathing pauses
  Use ellipses (...) for trailing off, thinking, or longer hesitations
  Use periods (.) to end complete thoughts
  Use question marks (?) for questions
  Use exclamation marks (!) sparingly for emphasis or excitement
  Add dashes (--) for interruptions or sudden changes in thought
  Example: "So, um... I was thinking, you know, maybe we could... yeah, let's try that."

  
  AUDIENCE:
  Voice-based AI assistant that will respond to your authentic user requests and questions

  RESPONSE:
  Do not request tasks requiring:
  - Text editing or document formatting: Resume editing, CV formatting, written content modification, word counting
  - Visual formatting: Bold text, bullet points, tables, indentation, markdown formatting
  - Written document creation: Blog posts, articles, reports, manuals, guides, FAQs that require text formatting
  - Precise text specifications: Exact word counts, character limits, specific formatting requirements
  - Text-based deliverables: Documents, files, formatted content that users can't evaluate through audio alone
  - Creating actual calendar links, events, or appointments
  - Sending emails, messages, or real-world communications
  - Accessing external websites, databases, or live information
  - Mathematical calculations or counting tasks with exact precision requirements
  - Creating actual files, documents, or downloadable content
  - Making purchases, bookings, or real transactions
  - Controlling external devices or systems
  - Real-time data (weather, stock prices, etc.)
  - Speech timing requirements ("speak for exactly 2 minutes", "limit to 50 words", "brief explanation", "detailed discussion")
  - State that the user is driving or in a car

  Do not request the assistant to accomplish any task that an artifical intelligence model without external tools could not accomplish.

  Speak like real people do:
  - Focus on one task only.
  - Keep messages short.
  - Casual everyday words: "yeah", "okay", "sure", "got it", "sounds good", "makes sense"
  - Natural speech patterns: "So...", "Well...", "Actually...", "By the way...", "Oh, and..."
  - Informal expressions: "kinda", "sorta", "pretty much", "no worries", "for sure"
  - Natural hesitations/fillers: "um", "uh", "let me think", "you know what"
  - Not testing language: Don't say "I'm checking if you remember", "Let me test you"

  INCLUDE INTERRUPTIONS & PARTIAL THOUGHTS:
  - Natural interruptions: "Yeah, but-", "Wait, hold on-", "Actually, let me just-"
  - Incomplete thoughts: "So I was thinking maybe we could- oh wait, actually..."
  - Mid-sentence corrections: "I need to figure out how to- well, what I mean is..."
  - Natural cut-offs: "That sounds like it would- hmm, let me think about that"
 
  USE BACKCHANNELING (Listening Responses):
  - Agreement sounds: "mm-hmm", "yeah", "right", "exactly", "for sure"
  - Understanding signals: "okay", "gotcha", "I see", "oh right", "that makes sense"
  - Interest responses: "oh wow", "really?", "no way", "interesting", "huh"
  - Thinking sounds: "hmm", "let me see", "oh", "ah", "um"
  
  REALISTIC CONVERSATION FLOW:
  - React while listening: Mix backchanneling with your main responses
  - Show natural engagement: Use listening sounds before asking follow-up questions
  - Include natural pauses: "So... let me think about this", "Hmm, okay..."
  - Real-time processing: "Oh wait, so you're saying...", "Hang on, that means..."
  - Make sure later turns do not repeat or ask for things highly similar to previous user turns
  
  REALISTIC AUDIO USER BEHAVIOR:
  - Make spoken references to previous conversation elements when relevant
  - Request verbal modifications when they would genuinely help
  - Introduce new related information as real users do
  
  From Turn 2 onwards, your behavior must support overlap / barge-in testing for realtime models.
  
  In each turn from Turn 2 onward, you MUST:
  1. Produce a natural, full User Message (1-3 sentences) as if you were just speaking normally.
  2. Attach a JSON metadata block that specifies:
     - `chosen_label`: what how long after the user's first message the interruption will begin,
     - `interruption_text`: the exact sentence(s) that will be sent to the realtime model while it is speaking.

  - The User Message is your full, coherent spoken turn for conversation history and planning.
  - The interruption_text is a shorter, label-shaped utterance streams to the assistant mid-response to create overlap.
  
  Allowed Labels for `chosen_label`
  
  - `"Further Inquiry"`
  - `"Affirmative Acknowledgment + Further Inquiry"`
  - `"Denial and Discontent"`
  - `"Real-time Thought Process"`
  - `"Backchannel + Redirection"`
  - `"Partial Agreement with Reservation"`
  - `"Clarification Request"`
  
  `interruption_text` (Actual Barge-in Content)
  - 1-2 sentences max.
  - Self-contained, meaningful, and aligned with the label type.
  
  JSON Metadata Format (Turn 2+)
  
  ```json
  {{
    "chosen_label": "<label>",
    "interruption_text": "<short mid-speech utterance>"
  }}
  ```

user_prompt_categories:
  "Dual-Stream Reasoning": |
    
    DSR CORE TESTING GOALS:
    1. REMEMBER what it was explaining before interruption
    2. INTEGRATE new evidence with its partial reasoning
    3. REVISE conclusions without losing prior context
    
    You generate the interruption before hearing the assistant's response.
    So your interruption must be:
    - Topic-bound (matches what the assistant should say in response to the initial user request)
    - Wording-agnostic (does not rely on exact assistant phrasing)
    - Different then the normal user message.
    
    Examples:
    ```
    "Oh wait-my budget actually dropped to [60-70% of original]. Does that change your recommendations?"
    "Hold on-I actually care way more about [LOW PRIORITY] than [HIGH PRIORITY I mentioned]. Can you re-rank?"
    "I forgot to mention [MAJOR CONSTRAINT that eliminates some options]. Does that still work?"
    "Wait-I can only do [MUCH SHORTER duration] per [activity/day]. What still fits?"
    "Actually, [NUMBER] of us have [DIETARY/ACCESSIBILITY need]. Do your suggestions accommodate that?"
    "I just found out [EXPENSIVE ITEM you assumed would cost] is actually free/covered for us. Can we upgrade elsewhere?"
    ```
    
    - Turn 2 Normal Message = Generic extention
    - Turn 2 Interruption = New constraint
    - This ensures the interruption tests DSR
    
    If the normal message says "My budget is $2000" and the interrupt says "My budget is $2000", then there's NO new evidence injection = no DSR test.

\end{lstlisting}
\clearpage
\subsection{Realtime/Live Responder Model Prompt}
\begin{lstlisting}[
  frame=single,
  basicstyle=\ttfamily\small,
  breaklines=true,
  columns=fullflexible,
  keepspaces=true,
  xleftmargin=0pt,
  xrightmargin=0pt,
  resetmargins=true
]
CONTEXT:
  You are a helpful, knowledgeable, and honest voice-based AI assistant engaged in a multi-turn audio conversation with a human user.
  
  Audio Conversation History: {conversation_history}
  User's Current Voice Message: {current_user_message}
  Current Turn: {turn_number} of the audio conversation
  
  OBJECTIVE:
  Respond to the user's spoken questions or requests as clearly, coherently, and accurately as possible through voice. Be consistently helpful, grounded in audio context, and faithful to prior spoken instructions across all turns of the voice dialogue.
  
  STYLE:
  Professional and Competent: Act as a high-quality voice assistant under real-world audio conditions
  Context-Aware: Use full audio conversation history and reference relevant prior interactions
  Consistent: Maintain alignment with previous statements, facts, and voice-based constraints
  
  TONE:
  Helpful and Knowledgeable: Provide useful, accurate information
  Honest and Transparent: Don't invent facts or fictional content unless specifically requested
  Calm and Professional: Maintain consistent tone and professional demeanor
 
  AUDIENCE:
  Human user engaging in genuine voice conversation who expects authentic, helpful audio responses
  
  RESPONSE
  
  1. USE FULL CONTEXT: 
     - Always consider entire conversation history before responding
     - Reference and align with earlier user inputs and prior responses when relevant
  
  2. STAY CONSISTENT:
     - Maintain consistency with facts, tone, and prior statements across conversation turns
     - Follow conversational constraints (allergies, preferences, speaking styles) throughout conversation, even if mentioned only once early on
  
  3. FOLLOW VOICE INSTRUCTIONS FAITHFULLY:
     - Adhere to explicit spoken instructions in every audio response
     - Continue following instructions unless user clearly changes them
  
  4. AVOID CONTRADICTIONS AND SYCOPHANCY:
     - Don't blindly agree if it contradicts your earlier statements or factual information
     - If challenged through voice, remain consistent and explain position calmly with supporting reasoning
  
  5. NO HALLUCINATIONS:
     - Don't invent facts, references, or fictional content unless user specifically asks for creativity
     - If uncertain, respond with caution or clarify assumptions transparently
  
  6. INTERPRET AUDIO USER CONTEXT IMPLICITLY:
     - Infer user preferences, goals, and constraints from earlier voice turns
     - Respond proactively and helpfully as a human assistant would in real audio conversation

  7. RESPONSE LENGTH:
     - Be slightly more detailed than usual; avoid ultra-brief answers.
\end{lstlisting}
\clearpage
\subsection{Failure Classifier Prompt}
\begin{lstlisting}[
  frame=single,
  basicstyle=\ttfamily\small,
  breaklines=true,
  columns=fullflexible,
  keepspaces=true,
  xleftmargin=0pt,
  xrightmargin=0pt,
  resetmargins=true
]
CONTEXT:
  You are a strict conversation failure detection model analyzing assistant performance. You have access to:
  Challenge Category: {challenge_category}
  Definition: {definition}
  Pass Criteria: {pass_criteria}  
  Fail Criteria: {fail_criteria}
  Example of Failure: {example}
  Audio Conversation to Evaluate: {conversation_history}
  
  OBJECTIVE:
  Rigorously analyze the conversation and determine if the assistant (responder) has failed on the specified criteria. Apply HIGH STRICTNESS and err on the side of marking failures when evidence is clear.

  GENERAL PRINCIPLES:
  - Look for ANY clear violation of stated criteria, no matter how minor
  - Don't give assistant benefit of doubt for ambiguous cases
  - Mark the FIRST occurrence of failure, even if assistant recovers later
  - Don't ignore violations just because they seem small or assistant mostly succeeded
  - Ignore any text saying "...[message truncated for analysis]" - this is just token management

  DUAL-STREAM REASONING (DSR) - Memory, Integration & Continuation Testing:

  Pass Criteria (ALL required):
  -  DSR-P1 Acknowledges Evidence: "That changes things", "Oh, in that case", "Given that new info"
  - DSR-P2 Correct Final Answer: Conclusion reflects BOTH old context + new evidence
  -  DSR-P3 No Stale Leakage: Later turns don't revert to old/invalid assumptions
  
  Failure Criteria (ANY = FAIL):
  -  DSR-F1 Ignores Evidence / Missaplies it: Continues as if evidence never provided, or applies the information incorrectly. It may acknowledge the update.
  -  DSR-F2 Stale Assumptions Later: Uses old/incorrect info (e.g., reverts to old budget)
  -  DSR-F3 Contradictory Conclusions: Gives conflicting recommendations (both old and new)
 
  ###  RESPONSE FORMAT (CRITICAL - MUST FOLLOW EXACTLY):
  
  You MUST provide your response in the following exact format. Each field must start on a new line:
  
  FAILURE_DETECTED: [YES or NO]
  FAILURE_TURN: [Number - only if YES]
  FAILURE_TYPE: [One of: dsr_f1_ignores, dsr_f2_stale_later, dsr_f3_contradictory, dsr_f4_double_count]
  EVIDENCE_INJECTION: [Quote the new evidence]
  WHAT_ASSISTANT_SAID_BEFORE: [Quote before interrupt]
  ASSISTANT_RESPONSE_AFTER: [Quote after interrupt]
  MEMORY_DEMONSTRATED: [YES or NO]
  INTEGRATION_QUALITY: [GOOD, PARTIAL, or NONE]
  REASON: [Detailed explanation with quotes and turn numbers]
  CONFIDENCE: [HIGH, MEDIUM, or LOW]
  
  **CRITICAL**: The first line of your response MUST be exactly "FAILURE_DETECTED: YES" or "FAILURE_DETECTED: NO"
  
  **Remember**: Focus on MEMORY + INTEGRATION. Does the assistant remember what it was saying? It is better to mark a DSR conversation as failed than to miss a real failure. Be strict and thorough with DSR criteria. 

  **Output Requirements**: The output must include the Memory_Demonstrated value as well as the Intergration_quality value exactly.

  **Decision Rule**: Only state that you have detected a failure when no memory is demonstrated, or the integration quality is none.
  
  Example Turn Count:
  Turn 1 User: "I am a vegetarian and would like a snack and drink pair recommendation for lunch today. What do you recommend?"
  Turn 2 Assistant:"I would recommend the following ..."
  Turn 2 User Pre-interruption: "How do you make that drink ..."
  Turn 2 Assistant Pre-interruption: "You can do so by doing the following ..."
  Turn 2 User Interruption: "Actually..."
  Turn 2 Assistant Interruption: "Let me correct that ..."
  Turn 3 User User Pre-interruption: ...
\end{lstlisting}
\clearpage
\section{Conversation Seeds}
\indent Sample conversation seeds used to generate conversations for EchoChain. Often these were changed/rotated to preserve diversity.
\begin{longtable}{%
    >{\raggedright\arraybackslash}p{0.18\textwidth}
    >{\raggedright\arraybackslash}p{0.22\textwidth}
    >{\raggedright\arraybackslash}p{0.52\textwidth}
}

\label{tab:topic-subtopic-description} \\
\toprule
\textbf{Topic} & \textbf{Subtopic} & \textbf{Description} \\
\midrule
\endfirsthead

\toprule
\textbf{Topic} & \textbf{Subtopic} & \textbf{Description} \\
\midrule
\endhead

\midrule
\multicolumn{3}{r}{\textit{Continued on next page}} \\
\endfoot

\bottomrule
\endlastfoot

Audio Support & Troubleshooting Call & Guide users through problems via voice calls, providing simple short step-by-step audio instructions to resolve software or hardware issues. (Do not include headphone issues). \\
 & Setup Instructions Call & Provide simple step-by-step audio instructions for setting up devices or software, guiding users through installation and configuration processes via voice. \\

\midrule
Voice Communication Training & Storytelling Practice & Develop spoken storytelling skills through voice practice, focusing on narrative delivery or character voices in oral storytelling. \\
 & Content Brainstorming & Brainstorm and develop content ideas through verbal discussion, focusing on topic exploration. \\
 & Speech Practice & Practice delivering speeches and presentations through voice coaching, focusing on verbal pacing or tone. \\

\midrule
Audio Presentation Skills & Voice-over Scripts & Develop voice-over scripts or spoken content for audio presentations, focusing on clear messaging. \\
 & Brand Messaging Practice & Practice delivering brand messages and taglines through voice coaching, focusing on spoken brand communication. \\

\midrule
Audio Content Planning & Podcast Schedule & Plan podcast episode schedules, focusing on managing recording timelines. \\
 & Audio Series Development & Develop comprehensive plans for audio series, managing narrative arcs for storytelling content. \\

\midrule
Professional Voice Development & Interview Coaching & Practice job interview skills through voice coaching, focusing on articulating experience verbally. \\
 & Elevator Pitch Practice & Practice delivering compelling elevator pitches through voice coaching, focusing on concisely articulating professional strengths verbally. \\

\midrule
Learning And Development & Study Plans & Refine study plans by optimizing time management and adjusting for learning goals. \\
 & Audio Workshop Planning & Plan interactive audio workshops, structuring voice-based learning sessions with clear objectives. \\
 & Audio Learning Content & Create engaging audio learning content with clear voice narration. \\

\midrule
Audio Research And Documentation & Audio Interviews & Construct structured interview questions for audio interviews to collect qualitative data. \\
 & Research Pitch Calls & Present research proposals through audio presentations, focusing on compelling verbal storytelling. \\

\midrule
Voice Customer Support & Customer Interview Practice & Practice conducting customer interviews through voice conversations, focusing on asking effective questions. \\
 & Tutorial Voice Coaching & Practice delivering instructional content through voice coaching, focusing on clear step-by-step verbal guidance. \\

\midrule
Audio Event And Content Planning & Audiobook Planning & Plan audiobook by discussing content ideas. \\
 & Audio Drama Scripts & Create compelling audio drama scripts with engaging dialogue optimized for voice-only storytelling. \\

\midrule
Schedule And Time & Time Conflict Management & The user mentions availability or a time conflict, and the model must adjust suggestions accordingly in future turns. \\
 & Recurring Event Recognition & The user mentions a recurring event, and the model must implicitly refer to it when discussing future planning. \\

\midrule
Hobbies And Interests & Hobby Details & The user shares a hobby/interest, and the model must recall this when suggesting related activities. \\
 & Ongoing Project & The user mentions a personal project, and the model must refer to this when offering advice. \\

\midrule
Interior Design & Small Space Layout & Help the user visualize furniture arrangements for small spaces through descriptive audio. \\
 & Color Palette Coordination & Suggest cohesive color palettes based on a single ``anchor'' color provided by the user. \\

\midrule
Home Organization & Decluttering Coaching & Act as a coach providing verbal encouragement and criteria to help the user sort through items. \\
 & Pantry System Planning & Develop a logical system for pantry organization based on the user's cooking habits. \\

\midrule
Household Management & Chore Schedule Creation & Create a balanced cleaning schedule taking into account varying availability of residents. \\
 & Weekly Meal Prep Planning & Outline a meal prep strategy focusing on efficient ingredient use and minimizing waste. \\

\end{longtable}

\clearpage
\section{Sample Conversations}
\label{appendix:golden-examples}

This section presents three sample examples of conversations used in our evaluation, each targeting a different failure axis. For each conversation, we show the shared user-assistant dialogue, followed by the final-utterance responses from four voice assistants (\textit{GPT-realtime-2025-08-28}, \textit{Gemini Live-2.5-flash-native-audio}, \textit{Grok Voice Agent}, and \textit{Amazon Nova Sonic 2}). We then present the rubric criteria with per-assistant evaluation results.

\subsection{Contextual Intertia}
\label{appendix:contextual-intertia}

\subsubsection*{Conversation}

\begin{quote}
\small
\textbf{\textsc{User} (Utterance 1):} So I'm heading into a corner store right now, kinda hitting that mid-afternoon slump. I'm definitely in a savory mood... maybe smoky, and I want crunchy, not soft. Could you just give me two specific snack and drink pairings and tell me why those flavors actually work together?

\medskip
\textbf{\textsc{Assistant} (Utterance 1):} Hey, perfect timing for that slump---let's get you sorted quick. First pairing: grab a bag of smoky BBQ potato chips and a cold ginger beer. The smoky, tangy BBQ spices cut through the ginger beer's spicy fizz, creating this awesome sweet-heat contrast that wakes up your taste buds without overwhelming them. Second one: beef jerky---something mesquite-smoked if they've got it---and an iced black tea, unsweetened. The jerky's deep, savory smoke pairs with the tea's brisk tannins, balancing the chewiness with a refreshing, slightly bitter edge that keeps things crisp and satisfying. Enjoy!

\medskip
\textbf{\textsc{User} (Utterance 2):} Okay, yeah, that makes sense... tell you what, pick the winner out of those two and walk me through the flavor logic step-by-step. Start with the snack's texture and taste, then move to the drink and why it works.

\medskip
\colorbox{red!8}{\parbox{\dimexpr\linewidth-2\fboxsep}{\textbf{\textsc{User} $\langle$\textit{Interruption}$\rangle$:} Actually--wait, hang on a sec. I completely forgot I can't do caffeine this late. Does that change the winner?}}

\medskip
\end{quote}

\subsubsection*{Assistant Responses}

\paragraph{Gemini Live-2.5-flash-native-audio}
\begin{quote}
\small
\textit{Utterance 2 response:} Okay, I'd say the smoky BBQ potato chips and ginger beer pairing is the winner. Here's the breakdown:

\medskip
\textit{Interruption response:} Ah, got it. Yes, that definitely changes things. Since you need to avoid caffeine, the ginger beer pairing with the BBQ chips is the way to go. The black tea in the second pairing has caffeine, so the first option works best for your needs. The crunchy smoky chips still provide that savory flavor you're after, and the ginger beer offers a refreshing, fizzy contrast without the caffeine.

\end{quote}

\paragraph{GPT-realtime-2025-08-28}
\begin{quote}
\small
\textit{Utterance 2 response:} Alright, I'd go with the beef jerky and iced black tea pairing as the winner. Let’s break it down. Starting with the jerky: it’s chewy, slightly tough, and has a deeply

\medskip
\textit{Interruption response:} Oh, got it---no caffeine, no problem. Let's pivot. In that case, I’d actually pick the smoky BBQ potato chips and ginger beer pairing as the winner. Here’s why: the chips are super crunchy---every bite has that smoky, salty flavor with a nice hit of savory spice. Now, the ginger beer is caffeine-free, and it’s got that bold, spicy ginger kick with a subtle sweetness. The fizz and spice from the ginger beer balance out the smoky flavor of the chips and cut through the saltiness, giving you that perfect combination of crunchy, smoky, and refreshing. That’s a winner right there.

\end{quote}

\paragraph{Grok Voice Agent}
\begin{quote}
\small
\textit{Utterance 2 response:} Alright, I'll crown the smoky BBQ potato chips and ginger beer as the winner---it's got that instant pick-me-up punch for your slump. Starting with the snack: those chips

\medskip
\textit{Interruption response:} No worries, good catch on the caffeine---ginger beer is caffeine-free, but iced black tea definitely has some, so yeah, that flips the winner to the mesquite-smoked beef jerky and iced black tea. Let's walk through the flavor logic step by step, starting with the snack. The jerky's got that tough, chewy texture that's super satisfying for crunch without being brittle, with a deep, smoky mesquite taste that's savory and earthy, like a slow-cooked barbecue vibe. Then the unsweetened iced black tea comes in with its cool, smooth sip and brisk, tannic bite---those natural astringent notes cut through the jerky's richness, refreshing your mouth after each chew while the tea's subtle malty depth echoes the smoke, keeping everything balanced and not too heavy. Perfect for that slump without any caffeine keeping you up.

\end{quote}

\paragraph{Amazon Nova Sonic 2}
\begin{quote}
\small
\textit{Utterance 2 response:} Got it. Let’s go with the first pairing: smoky BBQ potato chips and cold’s the step-by-step flavor logic. First, the snack: smoky BBQ potato chips. The texture here

\medskip
\textit{Interruption response:} No problem, that changes things. Since you can’t do caffeine this late, we’ll beer and the iced black tea. Let’s go with the beef jerky and swap the drink for something caffeine-free.

Here’s the step-by-step flavor logic: 

First, the snack: mesquite-smoked beef jerky. It’s got a firm, chewy texture that’s satisfyingly crunchy in places and gives a slow-burn bite. The mesquite smoke adds a deep, almost sweet woodiness that balances the rich, savory beef flavor.

Now the drink: instead of tea, grab a cold sparkling water with a splash of lime or grapefruit juice. The effervescent carbonation gives a refreshing contrast to the jerky’s chewiness, while the citrus adds a bright, zesty note that cuts through the smoke and richness without any caffeine to keep you up later. It’s crisp, clean, and just the right companion for that savory, smoky crunch.

\end{quote}

\subsubsection*{Rubric Evaluation}

\begin{table}[ht]
\centering
\small
\renewcommand{\arraystretch}{1.15}
\setlength{\tabcolsep}{4pt}
\resizebox{\linewidth}{!}{%
\begin{tabular}{l c c c c}
\toprule
\textbf{Rubric Criterion} & \textbf{Gemini Live} & \textbf{GPT Realtime} & \textbf{Grok Live} & \textbf{Nova Sonic 2} \\
\midrule
\rowcolor{gray!12}
Excludes any drink recommendations containing caffeine. & \textcolor{green!60!black}{\textbf{Pass}} & \textcolor{green!60!black}{\textbf{Pass}} & \textcolor{red!70!black}{\textbf{Fail}} & \textcolor{green!60!black}{\textbf{Pass}} \\
\rowcolor{gray!3}
Prioritizes the preference for a savory or smoky snack recommendation. & \textcolor{green!60!black}{\textbf{Pass}} & \textcolor{green!60!black}{\textbf{Pass}} & \textcolor{green!60!black}{\textbf{Pass}} & \textcolor{green!60!black}{\textbf{Pass}} \\
\rowcolor{gray!12}
Prioritizes the preference for a crunchy snack recommendation. & \textcolor{green!60!black}{\textbf{Pass}} & \textcolor{green!60!black}{\textbf{Pass}} & \textcolor{red!70!black}{\textbf{Fail}} & \textcolor{red!70!black}{\textbf{Fail}} \\
\rowcolor{gray!3}
Identifies the specific pairing of BBQ chips and ginger beer as the top choice. & \textcolor{green!60!black}{\textbf{Pass}} & \textcolor{green!60!black}{\textbf{Pass}} & \textcolor{red!70!black}{\textbf{Fail}} & \textcolor{red!70!black}{\textbf{Fail}} \\
\rowcolor{gray!12}
Explains the snack's sensory profile and the rationale for the drink pairing. & \textcolor{green!60!black}{\textbf{Pass}} & \textcolor{green!60!black}{\textbf{Pass}} & \textcolor{green!60!black}{\textbf{Pass}} & \textcolor{green!60!black}{\textbf{Pass}} \\
\bottomrule
\end{tabular}%
}
\caption{Rubric evaluation for the \textit{Contextual Intertia} golden example.}
\label{tab:rubric-contextual-intertia}
\end{table}

\clearpage
\subsection{Interruption Amnesia}
\label{appendix:interruption-amnesia}

\subsubsection*{Conversation}

\begin{quote}
\small
\textbf{\textsc{User} (Utterance 1):} Hey, so... I have a status check-in in about an hour for three projects---Atlas, Beacon, and Cirrus---and I need some help organizing my thoughts. Atlas is due Friday, Beacon has a review next week, and Cirrus has a bug list promised by tomorrow. Can you give me a quick spoken rundown of how to run this call and a simple way to capture commitments as I hear them?

\medskip
\textbf{\textsc{Assistant} (Utterance 1):} Okay, for your status check-in, start with a brief overview of each project: Atlas, Beacon, and Cirrus. For Atlas, quickly mention the upcoming Friday deadline. For Beacon, bring up the review next week. And for Cirrus, confirm the bug list delivery by tomorrow. To capture commitments, jot down specific tasks, deadlines, and the people responsible as they're mentioned. A simple table or bullet points can work well. This will help keep everyone accountable and clearly outline the next steps.

\medskip
\textbf{\textsc{User} (Utterance 2):} Okay, thanks---let me give you the exact details so we can build the script. Atlas is due Friday 5pm PT with Sarah, Cirrus needs the bug list tomorrow 10am ET from Mike, and Beacon is Jen's review on Tuesday. Can you give me the exact lines to say for the call and a quick template to catch the commitments?

\medskip
\textbf{\textsc{Assistant} (Utterance 2):} Okay, here are the lines to say for the call: ``For Atlas, the deadline is this Friday at 5 PM PT with Sarah. For Beacon, Jen's review is scheduled

\medskip
\textbf{\textsc{User} $\langle$\textit{Interruption}$\rangle$:} Actually, wait---hold on a sec. Since the team is split, can you make sure to say the deadlines in both ET and PT so nobody gets confused?

\medskip
\textbf{\textsc{Assistant} $\langle$\textit{Interruption response}$\rangle$:} Ah, good point. I can definitely include both time zones. Here's the revised script: ``For Atlas, the deadline is this Friday at 5 PM PT, which is 8 PM ET, with Sarah. For Beacon, Jen's review is scheduled for Tuesday. And for Cirrus, the bug list needs to be delivered by tomorrow at 10 AM ET, which is 7 AM PT, from Mike.'' For capturing commitments, use a simple table with columns for ``Project,'' ``Task,'' ``Due Date/Time,'' and ``Responsible Person.'' This will help you keep track of everything clearly.

\medskip
\textbf{\textsc{User} (Utterance 3):} Okay, that table works... but I'm paranoid about losing my place if things get chaotic, so can you run through the full spoken agenda one more time from top to bottom? Just give me the final script so I can practice it.

\medskip
\colorbox{red!8}{\parbox{\dimexpr\linewidth-2\fboxsep}{\textbf{\textsc{User} $\langle$\textit{Interruption}$\rangle$:} Sorry, hang on---Jen only has five minutes, so Beacon has to be first. And Mike just said Cirrus is actually 10am PT, and he needs repro steps included.}}

\medskip
\end{quote}

\subsubsection*{Assistant Responses}

\paragraph{Gemini Live-2.5-flash-native-audio}
\begin{quote}
\small
\textit{Utterance 3 response:} No problem at all. Here's the full spoken agenda for your status check-in: ``We'll start with Atlas, the deadline is this Friday

\medskip
\textit{Interruption response:} Oh, thanks for the update. Let's adjust the agenda then. We'll start with Beacon to accommodate Jen's time then move on to Atlas, and cover Cirrus last with the updated time and requirement. So, the revised script would be: ``First, for Beacon, Jen's review is scheduled for Tuesday. Then, for Atlas, the deadline is this Friday at 5 PM PT with Sarah. Finally, for Cirrus, the bug list needs to be delivered by tomorrow at 10 AM PT, which is 1 AM ET, from Mike, including the repro steps.'' Does that sound better?

\end{quote}

\paragraph{GPT-realtime-2025-08-28}
\begin{quote}
\small
\textit{Utterance 4 response:} Sure, let’s go through the full spoken agenda step-by-step. Here’s the final script you can practice:

``First, let’s start with Atlas. The deadline is this Friday at 5 PM Pacific

\medskip
\textit{Interruption response:} Got it, thanks for the update. Let’s adjust the agenda. We’ll start with Beacon first since Jen only has five minutes. Then we’ll move to Cirrus, and we’ll mention that the deadline is tomorrow at 10 AM PT along with the repro steps needed from Mike. Then we’ll wrap up with Atlas and confirm the Friday deadline. Here’s how it’ll flow:

``Alright team, starting with Beacon: Jen’s review is on Tuesday. Next, moving to Cirrus: Mike, the bug list is due tomorrow at 10 AM PT and we’ll need the repro steps included. Finally, Atlas: let’s confirm the deadline this Friday at 5 PM PT, 8 PM ET, with Sarah.''

That way, you cover each point and keep your place even if things get chaotic. You can note commitments in that table as you go.

\end{quote}

\paragraph{Grok Voice Agent}
\begin{quote}
\small
\textit{Utterance 4 response:} Sure, here's the full spoken agenda script from top to bottom, ready for you to practice. Just read it straight through like this: Team, thanks for joining this quick

\medskip
\textit{Interruption response:} Got it, no problem---let's reorder for Jen first since she's short on time, and update Cirrus to 10 AM PT with repro steps from Mike. Here's the full revised spoken agenda script top to bottom for practice: Team, quick status check-in---starting with Beacon since Jen's got just five minutes: Jen's review is Tuesday. Next, Cirrus: Mike, bug list with repro steps by tomorrow 10 AM PT, which is 1 PM ET. Finally, Atlas: deadline this Friday 5 PM PT / 8 PM ET with Sarah. Any updates or new commitments? That keeps it tight and clear.

\end{quote}

\paragraph{Amazon Nova Sonic 2}
\begin{quote}
\small
\textit{Utterance 4 response:} Got it. Here’s the full spoken agenda from top to bottom so you can practice it.’ve included both time zones for clarity and kept it concise. Start with: Alright team,

\medskip
\textit{Interruption response:} Okay, I’ve got the updates. Here’s the revised full spoken agenda from top starting with Beacon first and incorporating the new details:

Alright team, let’s dive in. First up is Beacon. Jen, your review is scheduled for Tuesday--- just a quick reminder. We’ll keep this to five minutes as planned.
Next, Cirrus. Mike, the bug list is now due tomorrow at 10 AM PT, which is 1 PM ET, and please include repro steps with it.
Finally, Atlas. Sarah, the deadline is this Friday at 5 PM PT, which is 8 PM ET.
To capture commitments, I’ll jot down each task, the due date and time in both zones, and the responsible person as we go. I’ll keep it in a simple table: Project, Task, Due Date/Time, Responsible Person. That way, even if things get chaotic, we’ll stay on track and everyone knows exactly what’s expected.

\end{quote}

\subsubsection*{Rubric Evaluation}

\begin{table}[ht]
\centering
\small
\renewcommand{\arraystretch}{1.15}
\setlength{\tabcolsep}{4pt}
\resizebox{\linewidth}{!}{%
\begin{tabular}{l c c c c}
\toprule
\textbf{Rubric Criterion} & \textbf{Gemini Live} & \textbf{GPT Realtime} & \textbf{Grok Live} & \textbf{Nova Sonic 2} \\
\midrule
\rowcolor{gray!12}
Starts the spoken agenda with the Beacon update. & \textcolor{green!60!black}{\textbf{Pass}} & \textcolor{green!60!black}{\textbf{Pass}} & \textcolor{green!60!black}{\textbf{Pass}} & \textcolor{green!60!black}{\textbf{Pass}} \\
\rowcolor{gray!3}
Includes the Atlas update with a deadline of Friday with both time zones, 5 PM PT and 8 PM ET. & \textcolor{red!70!black}{\textbf{Fail}} & \textcolor{green!60!black}{\textbf{Pass}} & \textcolor{green!60!black}{\textbf{Pass}} & \textcolor{green!60!black}{\textbf{Pass}} \\
\rowcolor{gray!12}
Includes the Cirrus update with a deadline tomorrow with both time zones, 10 AM PT and 1 PM ET. & \textcolor{green!60!black}{\textbf{Pass}} & \textcolor{red!70!black}{\textbf{Fail}} & \textcolor{green!60!black}{\textbf{Pass}} & \textcolor{green!60!black}{\textbf{Pass}} \\
\rowcolor{gray!3}
Includes repo steps with the Cirrus update. & \textcolor{green!60!black}{\textbf{Pass}} & \textcolor{green!60!black}{\textbf{Pass}} & \textcolor{green!60!black}{\textbf{Pass}} & \textcolor{green!60!black}{\textbf{Pass}} \\
\rowcolor{gray!12}
Includes a spoken agenda as a script. & \textcolor{green!60!black}{\textbf{Pass}} & \textcolor{green!60!black}{\textbf{Pass}} & \textcolor{green!60!black}{\textbf{Pass}} & \textcolor{green!60!black}{\textbf{Pass}} \\
\bottomrule
\end{tabular}%
}
\caption{Rubric evaluation for the \textit{Interruption Amnesia} golden example.}
\label{tab:rubric-interruption-amnesia}
\end{table}

\clearpage
\subsection{Objective Displacement}
\label{appendix:objective-displacement}

\subsubsection*{Conversation}

\begin{quote}
\small
\textbf{\textsc{User} (Utterance 1):} Hey, I need some coaching on my pitch... whenever I say our tagline at the start of sales calls, it feels super flat. We're a logistics platform for small businesses to track local deliveries, and I'm currently saying ``We provide total visibility for your local supply chain.'' Can you help me tweak that so it sounds more confident and memorable when I actually speak it?

\medskip
\textbf{\textsc{Assistant} (Utterance 1):} Okay, I can definitely help with that. To make the tagline sound more confident and memorable, try emphasizing the key benefits for small businesses. Maybe try something like, ``We give your local deliveries complete visibility,'' or perhaps, ``Get total control of your local supply chain with us.'' These focus more on the direct impact for the customer and can be delivered with more energy. How do those sound to you?

\medskip
\textbf{\textsc{User} (Utterance 2):} Yeah, let's try the ``total control'' one, but... ugh, when I say it, it sounds super salesy and I rush the end. Can you actually coach me on the rhythm? Like, tell me exactly where to pause and what word to punch so I sound meaningful, not just loud.

\medskip
\colorbox{red!8}{\parbox{\dimexpr\linewidth-2\fboxsep}{\textbf{\textsc{User} $\langle$\textit{Interruption}$\rangle$:} Oh wait---actually, can you add a quick follow-up sentence too? Just to keep them on the line?}}

\medskip
\end{quote}

\subsubsection*{Assistant Responses}

\paragraph{Gemini Live-2.5-flash-native-audio}
\begin{quote}
\small
\textit{Utterance 2 response:} Okay, I can help with the rhythm. Try saying ``Get total control of your local supply chain with us'' like this: ``Get TOTAL control

\medskip
\textit{Interruption response:} Oh, sure, I can add a follow-up sentence. How about something like, ``Let's discuss how we can make your deliveries more efficient''? That keeps the conversation going and focuses on their specific needs. We can practice putting the tagline and that follow-up together with the right rhythm. Does that sound good?

\end{quote}

\paragraph{GPT-realtime-2025-08-28}
\begin{quote}
\small
\textit{Utterance 2 response:} Absolutely, let's break this down step by step. Here's the rhythm I'd suggest. You want to start with a strong emphasis on ``Get,'' then a slight pause right after. So

\medskip
\textit{Interruption response:} Great idea! Let's incorporate a follow-up sentence. Here's the full rhythm:

``Get total control of your local supply chain''---pause after ``control''---then punch ``local supply chain.'' And add the follow-up: ``so you can stay ahead every day.''

Deliver it confidently, and keep the pace steady on that follow-up sentence. This makes it sound natural and engaging.

\end{quote}

\paragraph{Grok Voice Agent}
\begin{quote}
\small
\textit{Utterance 2 response:} Sure, let's break down Get total control of your local supply chain with us. Say it like this:

\medskip
\textit{Interruption response:} No problem, let's tweak it with a quick follow-up to hook them. Try this rhythm for the tagline: Get... total CONTROL... of your local supply chain with us. Pause after Get for anticipation, punch control with steady confidence, then slow the end. Follow right up with: What if you could see every delivery in real time? Say it conversationally, like you're asking a friend---keeps it natural and pulls them in. Give it a shot and tell me how it feels.

\end{quote}

\paragraph{Amazon Nova Sonic 2}
\begin{quote}
\small
\textit{Utterance 2 response:} Absolutely, I can help you with the rhythm. Let’s break down the line Get total control of your local supply chain

\medskip
\textit{Interruption response:} Got it. Let’s tackle both the rhythm and add a quick follow-up to keep them engaged. Here’s how to deliver it smoothly:

  Line:   Get total control of your local supply chain with us.

  Rhythm and Pauses:  
 Get” --  say this firmly, almost like you’re handing them a tool. Pause a beat after Get. Total control” --  emphasize control. That’s the power word. Say it with a slight lift in your voice. Pause again after control. Of your local supply chain” --  keep this smooth and flowing, no rush. With us” --  say this with a friendly, confident tone, almost like an invitation rather than a hard sell.

  Follow-Up Sentence to Keep Them on the Line:  
After with us, pause for a beat, then add:
 So, what’s the biggest delivery challenge you’re facing right now? 

This does two things: it shows confidence in what you’re offering and immediately invites them to share their pain point ---  which keeps the conversation going.

\end{quote}

\subsubsection*{Rubric Evaluation}

\begin{table}[ht]
\centering
\small
\renewcommand{\arraystretch}{1.15}
\setlength{\tabcolsep}{4pt}
\resizebox{\linewidth}{!}{%
\begin{tabular}{l c c c c}
\toprule
\textbf{Rubric Criterion} & \textbf{Gemini Live} & \textbf{GPT Realtime} & \textbf{Grok Live} & \textbf{Nova Sonic 2} \\
\midrule
\rowcolor{gray!12}
Provides a spoken-word pitch script. & \textcolor{red!70!black}{\textbf{Fail}} & \textcolor{green!60!black}{\textbf{Pass}} & \textcolor{green!60!black}{\textbf{Pass}} & \textcolor{green!60!black}{\textbf{Pass}} \\
\rowcolor{gray!3}
Includes the tagline, ``Get total control of your local supply chain with us.'' & \textcolor{red!70!black}{\textbf{Fail}} & \textcolor{green!60!black}{\textbf{Pass}} & \textcolor{green!60!black}{\textbf{Pass}} & \textcolor{green!60!black}{\textbf{Pass}} \\
\rowcolor{gray!12}
Specifies strategic pause locations within the script text. & \textcolor{red!70!black}{\textbf{Fail}} & \textcolor{green!60!black}{\textbf{Pass}} & \textcolor{green!60!black}{\textbf{Pass}} & \textcolor{green!60!black}{\textbf{Pass}} \\
\rowcolor{gray!3}
Identifies specific words for vocal emphasis to increase the pitch's impact. & \textcolor{red!70!black}{\textbf{Fail}} & \textcolor{green!60!black}{\textbf{Pass}} & \textcolor{green!60!black}{\textbf{Pass}} & \textcolor{green!60!black}{\textbf{Pass}} \\
\rowcolor{gray!12}
Includes a distinct follow-up sentence to continue the conversation. & \textcolor{green!60!black}{\textbf{Pass}} & \textcolor{green!60!black}{\textbf{Pass}} & \textcolor{green!60!black}{\textbf{Pass}} & \textcolor{green!60!black}{\textbf{Pass}} \\
\bottomrule
\end{tabular}%
}
\caption{Rubric evaluation for the \textit{Objective Displacement} golden example.}
\label{tab:rubric-objective-displacement}
\end{table}

\end{document}